\documentclass[11pt,a4paper]{article}
\usepackage{acl}
\usepackage{times}
\usepackage{latexsym}

\usepackage{graphicx}
\usepackage{amsmath}
\usepackage{mathrsfs}
\graphicspath{ {./graphs/} }
\usepackage{color, colortbl}
\usepackage{multirow}
\usepackage{stfloats}
\usepackage{lipsum} 

\usepackage{microtype}
\usepackage{amsfonts}
\usepackage{booktabs}
\usepackage{siunitx}

\title{Advancing Transformers' Capabilities in Commonsense Reasoning \\ (Technical Report)}

\author{Yu Zhou, Yunqiu Han, Hanyu Zhou, Yulun Wu\\
University of California, Los Angeles \\
\texttt{\{yu.zhou, yunqiu21, alexiazhou726, gloriawuyl\}@ucla.edu}}

\begin{document}
\maketitle

\begin{abstract}
Recent advances in general purpose pre-trained language models have shown great potential in commonsense reasoning. However, current works still perform poorly on standard commonsense reasoning benchmarks including the Com2Sense Dataset~\cite{singh-etal-2021-com2sense}. We argue that this is due to a disconnect with current cutting-edge machine learning methods. In this work, we aim to bridge the gap by introducing current ML-based methods to improve general purpose pre-trained language models in the task of commonsense reasoning. Specifically, we experiment with and systematically evaluate methods including knowledge transfer, model ensemble, and introducing an additional pairwise contrastive objective. Our best model outperforms the strongest previous works by $\sim$ 15\% absolute gains in Pairwise Accuracy and $\sim$ 8.7\% absolute gains in Standard Accuracy.~\footnote{Our code and data are publicly available for research purposes at \url{https://github.com/bryanzhou008/Advancing-Commonsense-Reasoning}}
\end{abstract}

\begin{figure}[t]
\begin{center}
\includegraphics[width=75mm,scale=0.5]{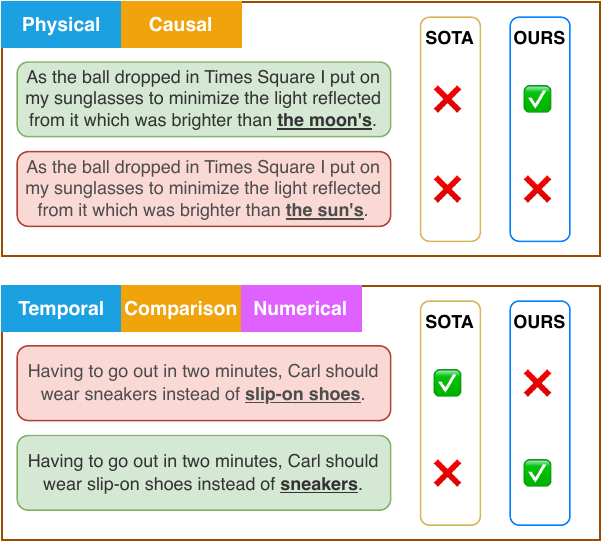}
\end{center}
\caption{Example data pairs from the Com2Sense test set along with model predictions. Here SOTA refers to \cite{Jung2022MaieuticPL} and OURS refers to Model 9 in Table~\ref{tbl:test}.}
\label{Fig: Qualitative}
\end{figure}

\section{Introduction}

Endowing NLP models with human-like commonsense knowledge has remained a challenge for decades \citep{sap-etal-2020-commonsense}. In 2021, researchers proposed Com2Sense~\citep{singh-etal-2021-com2sense}, a reliable and comprehensive commonsense reasoning benchmark with strict pairwise accuracy metrics. It consists of natural language sentence pairs labeled True/False based on whether they adhere to intuitive commonsense knowledge (Fig.\ref{Fig: Qualitative}). The central evaluation criteria: Pairwise Accuracy, required that the model predict correctly for both sequences to count as successful. 

Initial works on the dataset revealed that neither general purpose language models \cite{devlin-etal-2019-bert}, \cite{DBLP:journals/corr/abs-1907-11692}, \cite{Raffel2019ExploringTL}, etc. nor dedicated commonsense understanding models \cite{Khashabi2020UnifiedQACF}, \cite{2020unifiedqa} performed well on the dataset. All current models suffer from significant performance drops from Standard Accuracy to Pairwise Accuracy, displaying a huge discrepancy from human-like behaviour.

In this work, we examine possible methods of improving the performance of general purpose language models on the task of commonsense reasoning. Specifically, we study the effects of:
\begin{itemize}
    \item Knowledge Transfer from relevant datasets containing commonsense knowledge, including the SemEval-2020 Dataset\footnote{\url{https://alt.qcri.org/semeval2020/}} and the SQuAD2 QA Dataset~\cite{Rajpurkar2018KnowWY}
    \item Introducing a pairwise contrastive loss objective that forces models to distinguish commonsensical and non-commensensical statements
    \item Ensemble general purpose language models of different backbone architectures to study and compare their effects on overall performance.

\end{itemize}

\section{Methods}

\subsection{Knowledge Transfer}
\textbf{SemEval-2020 Dataset}
Similar to Com2Sense, \citet{wang-etal-2020-semeval} provided a commonsense-related dataset, SemEval-2020 Task 4, Commonsense Validation and Explanation(ComVE). Each instance included a pair of sentences, one of which makes sense while the other does not.

We hypothesize that using a language model pretrained on the SemEval dataset, we are able to achieve better performance than finetuning that model directly on Com2Sense. 
We trained the DeBERTaV3\textsubscript{large} model on SemEval dataset to get a checkpoint model. The parameters used for pretraining are: \emph{batch size = 48}, \emph{lr = 4e-5}, \emph{weight decay = 0.01}, \emph{adam eps = 1e-}6, trained for \emph{100 steps}.
We then finetuned the obtained model on the Com2Sense dataset with the same parameters as our best-performing model (Table~\ref{tbl:hyperparam}, Line 10) and compare the two models.

\noindent
\textbf{SQuAD2 QA Dataset}
We hypothesize that the language model will achieve higher performance after pretraining on question-answering datasets.
We compared the results from RoBERTa\textsubscript{base} and DeBERTaV3\textsubscript{large} with those from RoBERTa\textsubscript{base}-SQuAD2 and DeBERTaV3\textsubscript{large}-SQuAD2, respectively, by finetuning them on Com2Sense with the same parameters. We used the parameters in line 3 of Table~\ref{tbl:hyperparam} for RoBERTa\textsubscript{base} and parameters of our best model for DeBERTaV3\textsubscript{large}.

\subsection{Pairwise Contrastive Loss}
The Com2Sense dataset is complementary in nature. That is, for each statement, there is a complementary statement constructed with small perturbation on certain words making it concerning similar common sense concepts but with different (opposite) labels. This unique setting makes Com2Sense an ideal case for the use of contrastive learning.

We hypothesize that for the model to capture the semantic difference between commonsensical inputs vs their syntactically similar counterparts, it would be beneficial if we can push apart the hidden representation of each complementary input pair in the embedding space.

In practice, inspired by the InfoNCE Contrastive Loss by \citet{DBLP:journals/corr/abs-1807-03748}, we propose a Pairwise Contrastive Loss (PCL) function:

\begin{equation}
\mathscr{P}^{\mathrm{x}_{\mathrm{i}}, \mathrm{x}_{\mathrm{j}}}(\mathrm{W})=\frac{\mathrm{e}^{\operatorname{sim}\left(\operatorname{g}\left(\mathrm{x}_{\mathrm{i}}\right), \operatorname{g}\left(\mathrm{x}_{\mathrm{j}}\right)\right) / \tau}}{\sum_{\mathrm{k}=1, \mathrm{k} \neq \mathrm{i}}^{2 \mathrm{~N}} \mathrm{e}^{\operatorname{sim}\left(\operatorname{g}\left(\mathrm{x}_{\mathrm{i}}\right), \operatorname{g}\left(\mathrm{x}_{\mathrm{k}}\right)\right) / \tau}} \\
\end{equation}

Here for each complementary input sample pair $(\mathrm{x}_{\mathrm{i}},  \mathrm{x}_{\mathrm{j}})$ with embedding vectors $\operatorname{g}(\mathrm{x}_{\mathrm{i}}),  \operatorname{g}(\mathrm{x}_{\mathrm{j}})$, where $\operatorname{sim}\left(\operatorname{g}\left(\mathrm{x}_{\mathrm{i}}\right), \operatorname{g}\left(\mathrm{x}_{\mathrm{j}}\right)\right)$ is the dot product of the L2 normalised inputs and $\tau$ is the constant temperature parameter which we set to 0.5.

The total contrastive loss, $\mathcal{L}$, is defined as the arithmetic mean over all pairs in the batch of the cross entropy of their normalised similarities, i.e.

\begin{equation}
\mathcal{L}_{\text {total }}=-\frac{1}{\mathrm{~N}} \sum_{\mathrm{j}=1}^{\mathrm{N}} \log \mathscr{P}^{\mathrm{x}_{\mathrm{j}}, \mathrm{x}_{\mathrm{j}}}(\mathrm{W})
\end{equation}

\subsection{Model Ensemble and Rule-Based Perturbation}
Due to the complementary nature of the Com2Sense dataset, each input data pair should have one positive sample and one negative sample. With this fact in mind, we propose a posterior model ensemble pipeline that aims to reduce the number of Same-Output Pairs where two prediction labels are the same (either both positive or both negative). This method further helps our ensembled model distinguish between syntactically similar sentence pairs that represent different ideas.

In practice, we take $n$ finetuned models and rank them by their pairwise accuracy score on the dev set to represent our confidence in each model. Then we use the highest performing model as a base predictor to generate predictions on the test set, which would contain a number of Same-Output Pairs. For each Same-Output Pair, we move down the list of ranked models by confidence and generate their predictions. If the new model can differentiate the two samples (generating one positive and one negative), we then adopt the new model's prediction. In the end, we have an (ideally very small) number of test pairs that none of the models is able to differentiate between, in which case we randomly assign different prediction values to the pair.

\begin{table*}[t]
\makebox[\textwidth][c]{
\begin{tabular}{clccc} \toprule 
    {\textbf{Line}} & \multirow{2}{*}{\textbf{Model}} &   {\textbf{Pairwise}} & {\textbf{Standard}}\\
    {\textbf{No.}} & {}   & {\textbf{Acc \%}} & {\textbf{Acc \%}}\\\midrule
    1  & UnifiedQA-3B \cite{Khashabi2020UnifiedQACF}  & 51.26 & 71.31 \\ 
    2  & Maieutic Prompting \cite{Jung2022MaieuticPL}  & 68.70  & 75.00 \\ \hline \rowcolor{gray!20}
    3  & DeBERTaV3\textsubscript{large} (best tuned parameters)    & 63.40 & 77.87 \\ \rowcolor{gray!20}
    4  & DeBERTaV3\textsubscript{large} + KT  & 64.19 & 78.10 \\ \rowcolor{gray!20}
    5  & DeBERTaV3\textsubscript{large} + KT + CV  & 66.74 & 79.39 \\ \rowcolor{gray!20}
    6  & DeBERTaV3\textsubscript{large} + KT + CV + Contrastive + RP  & 82.07 & 82.07 \\ \rowcolor{gray!20}
    7  & DeBERTaV3\textsubscript{large} + KT + CV + Contrastive + Ensemble(5) + RP  & 82.62 & 82.62 \\ \rowcolor{gray!20}
    8  & DeBERTaV3\textsubscript{large} + KT + CV + Contrastive + Ensemble(8)  & 78.96 & 80.53 \\ \rowcolor{gray!20}
    9  & \textbf{DeBERTaV3\textsubscript{large} + KT + CV + Contrastive + Ensemble(8) + RP}  & \textbf{83.69} & \textbf{83.69}  \\ \hline
    10 & Human  & 95.00 &  96.50 \\ \bottomrule
\end{tabular}
}
\caption{Summary of our results on the Com2Sense test set: KT stands for Knowledge Transfer, CV stands for Cross Validation, RP stands for Rule-based Perturbation, Ensemble(5) stands for a 5-model ensemble between DeBERTaV3\textsubscript{large} and DeBERTaV3\textsubscript{base}, Ensemble(8) stands for an 8-model ensemble between DeBERTaV3\textsubscript{large}, DeBERTaV3\textsubscript{base}, and RoBERTa\textsubscript{base}. The best model and method is highlighted with bold texts. For fairness of comparison, all the above model performances are measured on the official Com2Sense test set.}~\label{tbl:test}
\end{table*}

\section{Results}

\subsection{Different Model Backbones}
To find the best model backbone architecture, we compare the results of BERT\textsubscript{base}, RoBERTa\textsubscript{base}, DeBERTa\textsubscript{base}, and DeBERTaV3\textsubscript{base} with the best finetuning parameters used by their respective authors. Our results show DeBERTaV3\textsubscript{base} to be the best structure with 48.74\% pairwise acc., while DeBERTa\textsubscript{base} and RoBERTa\textsubscript{base} have similar performance at $\sim$18\%. BERT\textsubscript{base} is the lowest performing model at $\sim$3\%

To find the best model size, we conduct multiple experiments with DeBERTaV3\textsubscript{base} and DeBERTaV3\textsubscript{large} under the best finetuning parameters used by \citet{DBLP:journals/corr/abs-2111-09543}. The results show that DeBERTaV3\textsubscript{large} reaches 68.34\% pairwise acc. while DeBERTaV3\textsubscript{base} reaches 52.76\%. This supports the hypothesis that larger models have stronger common sense reasoning ability.


After choosing the best performing model DeBERTaV3\textsubscript{large} as our base model, we perform hyperparameter tuning on model parameters including: batch size (equivalent batch size after gradient accumulation), learning rate, and warmup steps. In each case, we fix all other parameters and test the effect of different values for the parameter under investigation. The testing results are documented in Table~\ref{tbl:hyperparam} lines 9-14, and the best set of parameters is highlighted in line 10.

\subsection{Knowledge Transfer}

\subsubsection{SemEval-2020 Dataset} As shown in the following table, the model with transferred knowledge from SemEval-2020 performs better than the best model directly applied to Com2Sense, with a 0.364\% improvement on pairwise accuracy.

\noindent
\makebox[\linewidth][c]{
\begin{tabular}{lcc} \toprule 
    \multirow{2}{*}{\textbf{Model}} & {\textbf{Pairwise}} & {\textbf{F1-}}\\ 
     & {\textbf{Acc \%}} & {\textbf{Score}} \\\midrule
    best DeBERTaV3\textsubscript{large} & 68.34 & 0.8103 \\
    SemEval-pretrained & 68.84 & 0.8139 \\
\bottomrule
\end{tabular}
}

\hspace{0.3cm}

\subsubsection{SQuAD2 QA Dataset} 
From the following table, we can see that models pretrained on question-answering data did not perform as well as those that have not. This can be because question-answering is a vastly different task than binary classification.

\hspace{0.3cm}

\noindent
\makebox[\linewidth]{
\begin{tabular}{lcc} \toprule 
    \multirow{2}{*}{\textbf{Model}} & {\textbf{Pairwise}} & {\textbf{F1-}} \\
     & {\textbf{Acc \%}} & {\textbf{Score}} \\\midrule
    RoBERTa\textsubscript{base}  & 18.84 & 0.546 \\
    RoBERTa\textsubscript{base}-SQuAD2 & 13.56 & 0.554 \\
    DeBERTaV3\textsubscript{large}  & 68.34 & 0.810 \\
    DeBERTaV3\textsubscript{large}-SQuAD2 & 65.83 & 0.789 \\
\bottomrule
\label{tab: squad2}
\end{tabular}
}


\subsection{Contrastive Learning and Random Perturbation}
From Table~\ref{tbl:test} lines 5-8, we observe that in practice contrastive learning together with the Random Perturbation helped to improve test performance by ~16\%. In this case, we count that random perturbation changed 371 out of 2790 pairs in the test set. While this can have a maximum influence of 13.29\% if all changed pairs turn out to be correct, since it is a purely random perturbation, on average it should have improved pairwise accuracy by 6.65\%.

After removing the benefits of Random Perturbation, we conclude that Contrastive Learning yields an improvement of 8.77\% on average, 2.04\% in the worst case. The improvement can be attributed to the fact that the Com2Sense dataset comes in a natural contrastive fashion, with uniform true/false pairs that need to be differentiated from each other.

\subsection{Model Ensemble}
From Table~\ref{tbl:test} lines 6-8, we observe that in practice model ensemble as a post-processing technique helps the model perform better compared to straight-through Random Perturbation, likely because Random Perturbation only has a 50\% chance of correctly predicting a pair while models used in the ensemble have a much higher accuracy.

In addition, the ensemble among DeBERTaV3\textsubscript{large}, DeBERTaV3\textsubscript{base}, and RoBERTa\textsubscript{base} models outperforms the ensemble between DeBERTaV3\textsubscript{large} and DeBERTaV3\textsubscript{base} models by 1.07\% pairwise acc. This results supports the common understanding that diversity in model structures is beneficial for the ensemble.

\section{Discussion}
\subsection{Results Analysis}
We make use of the domain, scenario, and numeracy dimensions of Com2Sense, take the best-performing model of BERT\textsubscript{base}, DeBERTaV3\textsubscript{base}, DeBERTaV3\textsubscript{large}, and SemEval-pretrained DeBERTV3\textsubscript{large}, and then calculate each model's pairwise accuracy on Com2Sense dev set in every possible combination of the three dimensions. 

\begin{figure*}[h]
    \centering
    \includegraphics[width=0.716\linewidth]{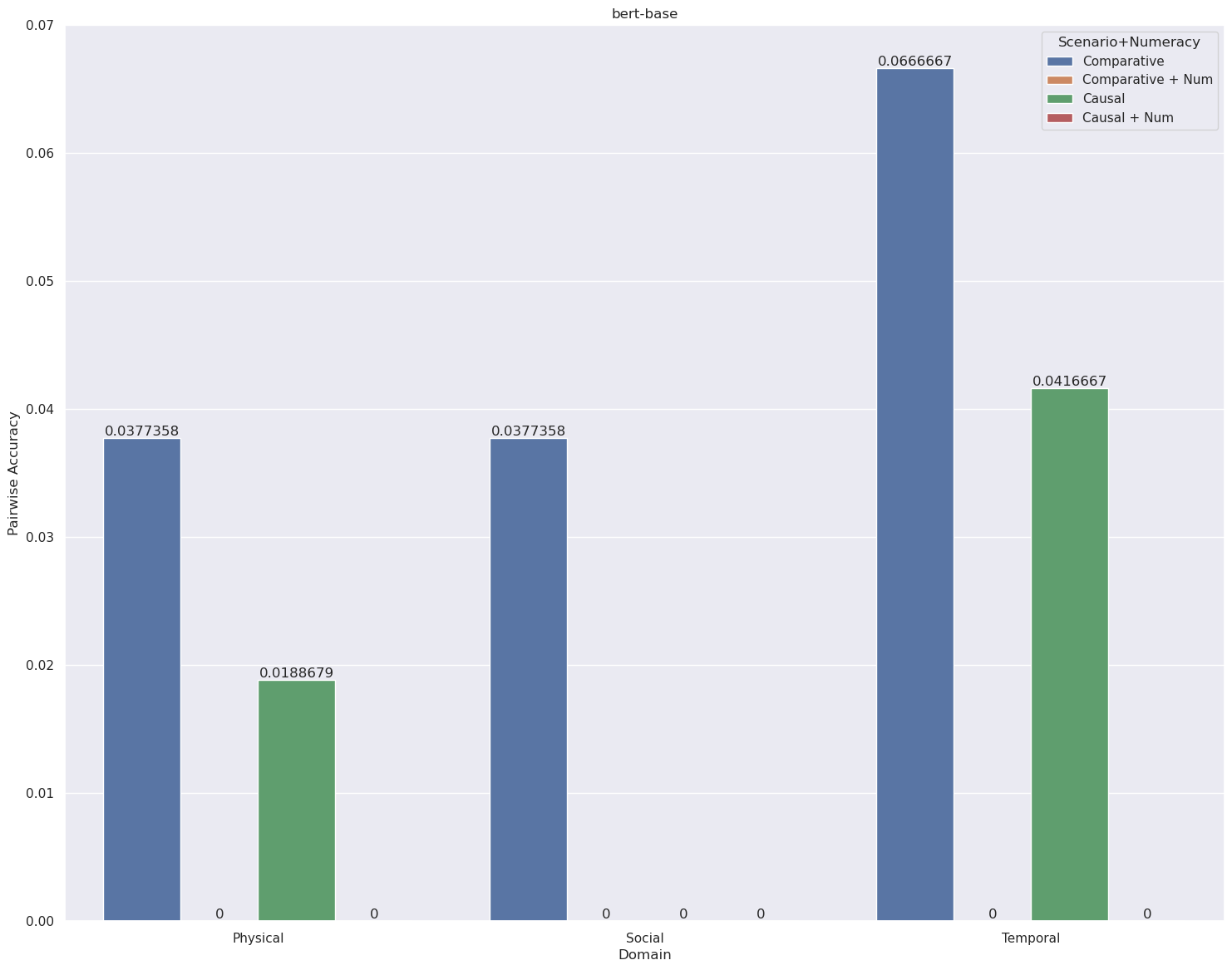}
    \includegraphics[width=0.716\linewidth]{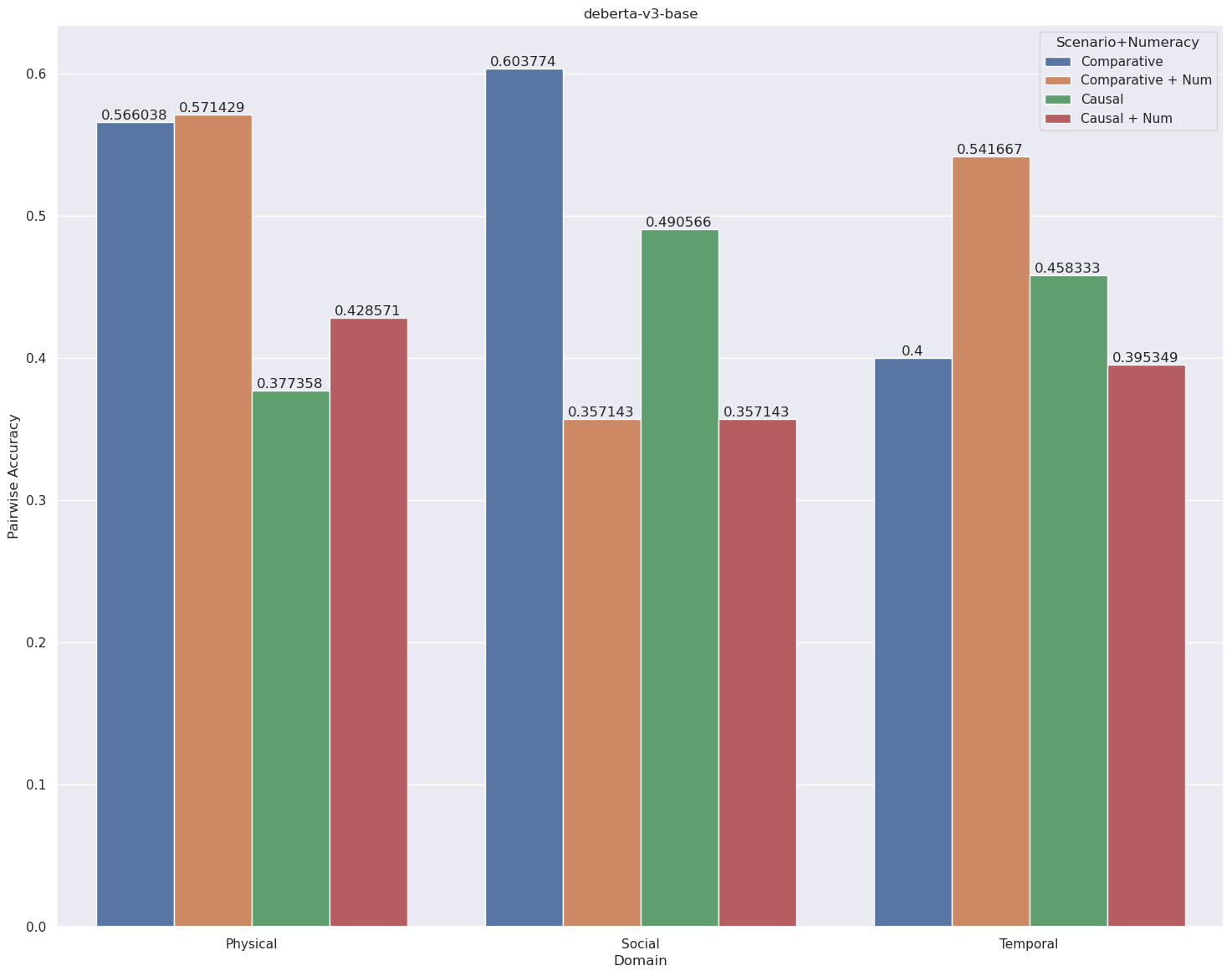}
    \caption{The top graph shows the pairwise accuracy of different dimension combinations for BERT\textsubscript{base}, and the graph below shows the pairwise accuracy of different dimension combinations for DeBERTaV3\textsubscript{base}.}
    \label{fig:base}
\end{figure*}

In general, BERT\textsubscript{base} gives the lowest pairwise accuracy, as shown in Table \ref{tbl:hyperparam}. The top graph in Figure \ref{fig:base} further reveals that the model correctly predicts none of the numeracy data. Comparatively, it gives better predictions on sentences with comparison than with causal relationship, and yields higher pairwise acc. on temporal sentences than physical and, lastly, social.

DeBERTaV3\textsubscript{base} gives boosted pairwise accuracy. From the bottom graph in Figure \ref{fig:base}, it performs better on comparative data than causal for all domains. We get slightly better results with numeracy than without numeracy for the physical domain, but in reverse for the social domain. The pattern for the temporal domain is more mixed: data with numeracy information has higher pairwise accuracy than comparisons but decreases for causal scenarios.

\begin{figure*}[h]
    \centering
    \includegraphics[width=0.716\linewidth]{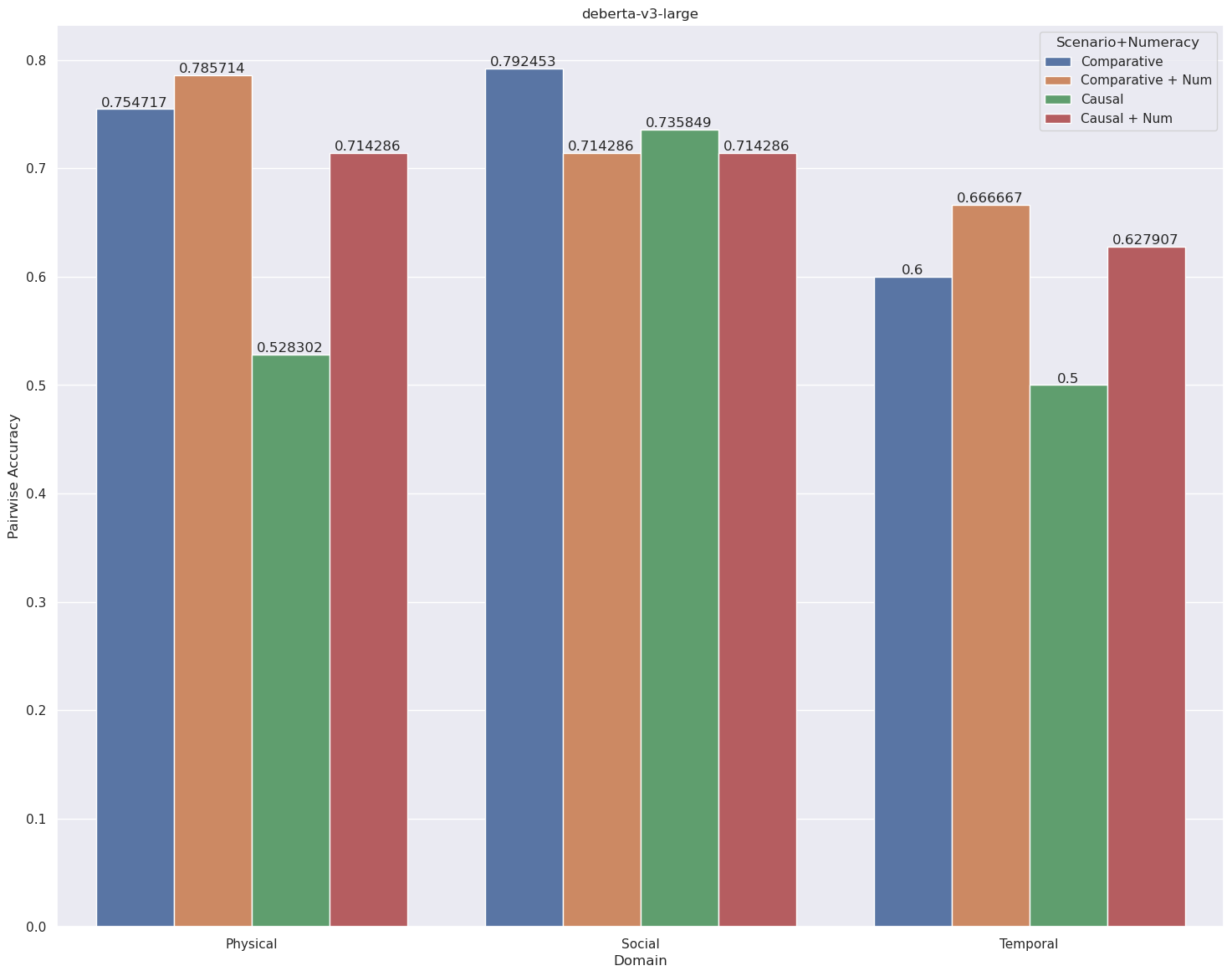}
    \includegraphics[width=0.716\linewidth]{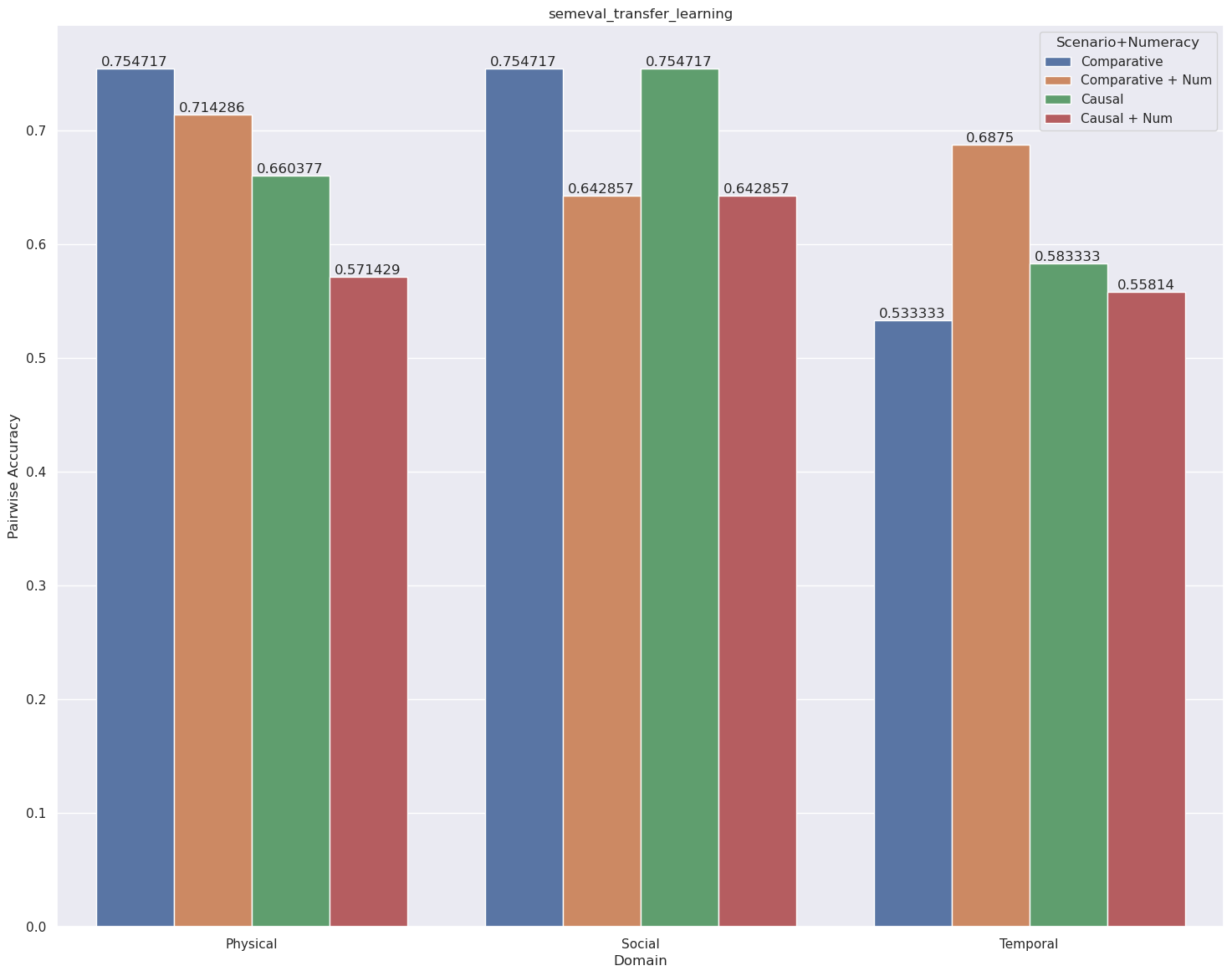}
    \caption{The top graph shows the pairwise accuracy of different dimension combinations for BERT\textsubscript{large}, and the graph below shows the pairwise accuracy of different dimension combinations for DeBERTaV3\textsubscript{large} pretrained on SemEval dataset.}
    \label{fig:large}
\end{figure*}

DeBERTaV3\textsubscript{large} improves the pairwise accuracy for all categories, but in particular more salient for social domain and numeracy data, as shown in Figure \ref{fig:large} top graph. It mostly preserves the pattern of DeBERTaV3\textsubscript{base}, while performing better on data with numeracy information for temporal, causal sentences.

DeBERTV3\textsubscript{large} pretrained on SemEval dataset (Figure \ref{fig:large} bottom graph) generally performs better in social domain than physical, worst in temporal; it also improves on data without numeracy. We do observe, though, the exceptionally higher performance on temporal, comparative, and numeric sentences, probably affected by knowledge transferred from SemEval. The pretraining might have also improved DeBERTV3\textsubscript{large}'s ability to learn causal reasoning as its pairwise accuracy increases for causal scenarios.

\subsection{Limitations}
The following are areas that we could improve during the stages of training and finetuning. Firstly, due to different hardware limitations on each of our virtual machines, we were not able to maintain consistent per-GPU batch size when training models throughout the project. While some of the trials used 6-instance batches over 8 accumulation steps, others were only able to use 4-instance batches over 12 accumulation steps. Such inconsistency in parameters could have impacted our final results. Secondly, due to time constraints, we only tested a limited range of hyperparameters which was not guaranteed to be the global optimum.

In terms of method generality, one limiting factor is our use of the Pairwise Contrastive Loss. That specific loss is reliant upon the paired structure of the Com2Sense training set. With information of paired commonsensical/non-commonsensical inputs, the specific loss is able to better encode information and instruct the model at hand compared to a simple cross-entropy loss. However, this construction is not generalizable to other datasets without such paired-input training data, thus it is another limitation of our method.

\section{Conclusion}
In this on-going research project, we have experimented with various methods to improve general purpose language models on the commonsense learning and reasoning task benchmarked by Com2Sense. We showed that: knowledge transfer from existing commonsense datasets; pairwise contrastive learning from commonsensical statements and their perturbed counterparts; and ensembling models with heterogeneous backbones yielded the greatest overall performance gains among other methods. Experiments applying the methods demonstrate substantial improvements in all metrics over current SOTA works in the field.

\section{Acknowledgements}

This project was completed during CS188-NLP at UCLA, with support from current and previous course instructors Zi-Yi Dou, Te-Lin Wu, and Prof. Nanyun Peng. This work is supported by the Machine Common Sense (MCS) program under Cooperative Agreement N66001-19-2-4032 with the US Defense Advanced Research Projects Agency (DARPA). The views and the conclusions of this work are those of the authors and do not reflect the official policy or position of DARPA.


\bibliographystyle{acl_natbib}
\bibliography{acl2023.bib}

\begin{thebibliography}{12}
\expandafter\ifx\csname natexlab\endcsname\relax\def\natexlab#1{#1}\fi

\bibitem[{Devlin et~al.(2019)Devlin, Chang, Lee, and
  Toutanova}]{devlin-etal-2019-bert}
Jacob Devlin, Ming-Wei Chang, Kenton Lee, and Kristina Toutanova. 2019.
\newblock \href {https://doi.org/10.18653/v1/N19-1423} {{BERT}: Pre-training of
  deep bidirectional transformers for language understanding}.
\newblock In \emph{Proceedings of the 2019 Conference of the North {A}merican
  Chapter of the Association for Computational Linguistics: Human Language
  Technologies, Volume 1 (Long and Short Papers)}, pages 4171--4186,
  Minneapolis, Minnesota. Association for Computational Linguistics.

\bibitem[{He et~al.(2021)He, Gao, and Chen}]{DBLP:journals/corr/abs-2111-09543}
Pengcheng He, Jianfeng Gao, and Weizhu Chen. 2021.
\newblock \href {http://arxiv.org/abs/2111.09543} {Debertav3: Improving deberta
  using electra-style pre-training with gradient-disentangled embedding
  sharing}.
\newblock \emph{CoRR}, abs/2111.09543.

\bibitem[{Jung et~al.(2022)Jung, Qin, Welleck, Brahman, Bhagavatula, Bras, and
  Choi}]{Jung2022MaieuticPL}
Jaehun Jung, Lianhui Qin, Sean Welleck, Faeze Brahman, Chandra Bhagavatula,
  Ronan~Le Bras, and Yejin Choi. 2022.
\newblock Maieutic prompting: Logically consistent reasoning with recursive
  explanations.
\newblock In \emph{Conference on Empirical Methods in Natural Language
  Processing}.

\bibitem[{Khashabi et~al.(2020{\natexlab{a}})Khashabi, Min, Khot, Sabhwaral,
  Tafjord, Clark, and Hajishirzi}]{2020unifiedqa}
D.~Khashabi, S.~Min, T.~Khot, A.~Sabhwaral, O.~Tafjord, P.~Clark, and
  H.~Hajishirzi. 2020{\natexlab{a}}.
\newblock Unifiedqa: Crossing format boundaries with a single qa system.

\bibitem[{Khashabi et~al.(2020{\natexlab{b}})Khashabi, Min, Khot, Sabharwal,
  Tafjord, Clark, and Hajishirzi}]{Khashabi2020UnifiedQACF}
Daniel Khashabi, Sewon Min, Tushar Khot, Ashish Sabharwal, Oyvind Tafjord,
  Peter Clark, and Hannaneh Hajishirzi. 2020{\natexlab{b}}.
\newblock Unifiedqa: Crossing format boundaries with a single qa system.
\newblock In \emph{Findings}.

\bibitem[{Liu et~al.(2019)Liu, Ott, Goyal, Du, Joshi, Chen, Levy, Lewis,
  Zettlemoyer, and Stoyanov}]{DBLP:journals/corr/abs-1907-11692}
Yinhan Liu, Myle Ott, Naman Goyal, Jingfei Du, Mandar Joshi, Danqi Chen, Omer
  Levy, Mike Lewis, Luke Zettlemoyer, and Veselin Stoyanov. 2019.
\newblock \href {http://arxiv.org/abs/1907.11692} {Roberta: {A} robustly
  optimized {BERT} pretraining approach}.
\newblock \emph{CoRR}, abs/1907.11692.

\bibitem[{Raffel et~al.(2019)Raffel, Shazeer, Roberts, Lee, Narang, Matena,
  Zhou, Li, and Liu}]{Raffel2019ExploringTL}
Colin Raffel, Noam~M. Shazeer, Adam Roberts, Katherine Lee, Sharan Narang,
  Michael Matena, Yanqi Zhou, Wei Li, and Peter~J. Liu. 2019.
\newblock Exploring the limits of transfer learning with a unified text-to-text
  transformer.
\newblock \emph{ArXiv}, abs/1910.10683.

\bibitem[{Rajpurkar et~al.(2018)Rajpurkar, Jia, and
  Liang}]{Rajpurkar2018KnowWY}
Pranav Rajpurkar, Robin Jia, and Percy Liang. 2018.
\newblock Know what you don’t know: Unanswerable questions for squad.
\newblock In \emph{Annual Meeting of the Association for Computational
  Linguistics}.

\bibitem[{Sap et~al.(2020)Sap, Shwartz, Bosselut, Choi, and
  Roth}]{sap-etal-2020-commonsense}
Maarten Sap, Vered Shwartz, Antoine Bosselut, Yejin Choi, and Dan Roth. 2020.
\newblock \href {https://doi.org/10.18653/v1/2020.acl-tutorials.7} {Commonsense
  reasoning for natural language processing}.
\newblock In \emph{Proceedings of the 58th Annual Meeting of the Association
  for Computational Linguistics: Tutorial Abstracts}, pages 27--33, Online.
  Association for Computational Linguistics.

\bibitem[{Singh et~al.(2021)Singh, Wen, Hou, Alipoormolabashi, Wu, Ma, and
  Peng}]{singh-etal-2021-com2sense}
Shikhar Singh, Nuan Wen, Yu~Hou, Pegah Alipoormolabashi, Te-lin Wu, Xuezhe Ma,
  and Nanyun Peng. 2021.
\newblock \href {https://doi.org/10.18653/v1/2021.findings-acl.78}
  {{COM}2{SENSE}: A commonsense reasoning benchmark with complementary
  sentences}.
\newblock In \emph{Findings of the Association for Computational Linguistics:
  ACL-IJCNLP 2021}, pages 883--898, Online. Association for Computational
  Linguistics.

\bibitem[{van~den Oord et~al.(2018)van~den Oord, Li, and
  Vinyals}]{DBLP:journals/corr/abs-1807-03748}
A{\"{a}}ron van~den Oord, Yazhe Li, and Oriol Vinyals. 2018.
\newblock \href {http://arxiv.org/abs/1807.03748} {Representation learning with
  contrastive predictive coding}.
\newblock \emph{CoRR}, abs/1807.03748.

\bibitem[{Wang et~al.(2020)Wang, Liang, Jin, Wang, Zhu, and
  Zhang}]{wang-etal-2020-semeval}
Cunxiang Wang, Shuailong Liang, Yili Jin, Yilong Wang, Xiaodan Zhu, and Yue
  Zhang. 2020.
\newblock {S}em{E}val-2020 task 4: Commonsense validation and explanation.
\newblock In \emph{Proceedings of The 14th International Workshop on Semantic
  Evaluation}. Association for Computational Linguistics.

\end{thebibliography}

\clearpage
\onecolumn

\section{Appendix}
\subsection{Performance Breakdown}

\clearpage
\subsection{Table of Best Parameters}

\begin{table*}[h]
\makebox[\textwidth][c]{
\begin{tabular}{clcccccccc} \toprule 
    {\textbf{Line}} & {\textbf{Model}} & {\textbf{best}} & {\textbf{batch}} & {\textbf{lr}} & {\textbf{weight}} & {\textbf{adam}} & {\textbf{warmup}} & {\textbf{Pairwise}} & {\textbf{F1 score}}\\
    {\textbf{No.}} & {} & {\textbf{step}} & {\textbf{size}} & {} & {\textbf{decay}} & {\textbf{$\epsilon$}} & {\textbf{step}} & {\textbf{Acc \%}} & {}\\\midrule
    1 & BERT\textsubscript{base} & 1020 & 32 & 1e-5 & 0 & 1e-8 & 0 & 3.01 & 0.4073 \\
    2 & BERT\textsubscript{large} & 60 & 64  & 5e-5 & 0  & 1e-8 & 0 & 2.51  & 0.3720  \\ \midrule
    3 & RoBERTa\textsubscript{base} & 1040 & 64  & 1e-5 & 0.01  & 1e-8 & 0 & 18.84  & 0.5463 \\ \midrule
    4 & DeBERTa\textsubscript{base} & 6000 & 32   & 1e-5 & 0   & 1e-8 & 500 & 17.84 & 0.5302 \\
    5 & DeBERTaV3\textsubscript{base} & 4500 & 48  & 1e-5 & 0.01  & 1e-6 & 500 & 48.74   & 0.7145  \\
    6 & DeBERTaV3\textsubscript{base} & 1500 & 48  & 3e-5 & 0.01  & 1e-6 & 100 & 52.76   & 0.7219  \\
    7 & DeBERTaV3\textsubscript{base} & 2500 & 48  & 3e-5 & 0.01  & 1e-6 & 500 & 49.00  & 0.7057  \\
    8 & DeBERTaV3\textsubscript{base} & 1000 & 48  & 9e-6 & 0.01  & 1e-6 & 500 & 45.48  & 0.6767  \\
    9 & DeBERTaV3\textsubscript{large} & 750 & 64  & 9e-6 & 0.01  & 1e-6 & 500 & 67.84  & 0.8090 \\
    10 & \textbf{DeBERTaV3\textsubscript{large}} & \textbf{1900} & \textbf{48}  & \textbf{9e-6} & \textbf{0.01}   & \textbf{1e-6} & \textbf{500} & \textbf{68.34}  & \textbf{0.8103}  \\
    11 & DeBERTaV3\textsubscript{large} & 1000 & 48  & 8.5e-6 & 0.01   & 1e-6 & 500 & 67.34  & 0.8111  \\
    12 & DeBERTaV3\textsubscript{large} & 450 & 48  & 9.5e-6 & 0.01  & 1e-6 & 500 & 66.33  & 0.7990 \\
    13 & DeBERTaV3\textsubscript{large} & 1000 & 48  & 9e-6 & 0.01   & 1e-6 & 300 & 67.59  & 0.8059  \\
    14 & DeBERTaV3\textsubscript{large} & 1400 & 48  & 9e-6 & 0.01  & 1e-6 & 750 & 66.58   & 0.8029 \\ \bottomrule
\end{tabular}
}
\caption{Summary of hyper-parameter tuning with results calculated on the dev dataset, the experiments are focused on finding the best model backbone, model size and ideal values for hyper-parameters. The best performing model and ideal hyper-parameter group is highlighted in bold.}~\label{tbl:hyperparam}
\end{table*}

\twocolumn
\subsection{Cross Validation Details}
The training dataset contains 797 pairs of examples and the development set has 398 pairs. We hypothesize that leveraging both datasets for training would yield a more generalized model with a higher level of reliability. We thus employed $k$-fold cross validation on our best DeBERTaV3\textsubscript{large} model and tested for $k = 2$ and $k = 5$.

As shown in Table~\ref{tbl:test} lines 3-5, cross validation helped to improve test performance by ~2.5\% empirically by incorporating the dev set for finetuning, which added around 50\% more training data. Consequently, the training time escalates with the number of folds: 2-fold cross validation took around 20 hours to train and 5-fold took more than 50 hours. 

\end{document}